%% file: spo.tex
\documentclass[10pt,twocolumn,letterpaper]{article}

\usepackage[pagenumbers]{cvpr} 


\usepackage{graphicx}
\usepackage{xcolor}
\usepackage{booktabs}
\usepackage{colortbl}
\usepackage{enumitem}
\usepackage{CJKutf8}
\usepackage{pifont}

\newcommand{\redtext}[1]{{\textcolor{red}{#1}}}

\definecolor{cvprblue}{rgb}{0.21,0.49,0.74}
\usepackage[pagebackref,breaklinks,colorlinks,allcolors=cvprblue]{hyperref}

\title{Aesthetic Post-Training Diffusion Models \\from Generic Preferences with Step-by-step Preference Optimization}

\author{
    \small
    Zhanhao Liang$^{1\dagger}$,
    Yuhui Yuan$^{5}$,
    Shuyang Gu$^{5}$,
    Bohan Chen$^{2\dagger}$,
    Tiankai Hang$^{3\dagger}$,
    Mingxi Cheng$^{4}$,
    Ji Li$^{4}$, 
    Liang Zheng$^{1}$ \\ 
    \small \texttt{zhanhao.liang@anu.edu.au, \{yuyua,shuyanggu\}@microsoft.com, liang.zheng@anu.edu.au} \\
    \small
    $^1$The Australian National University~~~
    $^2$University of Liverpool~~~
    $^3$Southeast University~~~ 
    $^4$Microsoft~~~
    $^5$Microsoft Research Asia~~~
}

\newcommand{\tablestyle}[2]{\setlength{\tabcolsep}{#1}\renewcommand{\arraystretch}{#2}\centering\footnotesize}
\newlength\savewidth\newcommand\shline{\noalign{\global\savewidth\arrayrulewidth
\global\arrayrulewidth 1pt}\hline\noalign{\global\arrayrulewidth\savewidth}}

\begin{document}
\begin{CJK}{UTF8}{gbsn}

\twocolumn[{
\maketitle
\begin{center}
    \vskip -0.2in
    \centering
    \includegraphics[width=1\textwidth]{imgs/intro_vis.pdf}
   \vskip -0.05in
    \captionof{figure}{Problem illustration of existing DPO methods. 
    We show two denoising trajectories of a preferred image (a) and a dispreferred image (b) generated from prompt ``A {cat} jumps on a {dog}''. (1) \textit{Disagreement between generic preferences and aesthetic preference}. While (a) is preferred due to its better layout, its details are poorer than (b). Red boxes show the erroneous fusion of the cat's leg and the dog's body. (2) \textit{Large sample differences between trajectories.} This makes it difficult to identify subtle aesthetic-related differences at each step.} 
    \label{fig:trajc_problem_vis}
\end{center}
}]

\begingroup
\renewcommand\thefootnote{}
\footnotetext{\textsuperscript{$\dagger$} Interns at Microsoft Research Asia.}
\endgroup

\begin{abstract}
Generating visually appealing images is fundamental to modern text-to-image generation models. A potential solution to better aesthetics is direct preference optimization (DPO), which has been applied to diffusion models to improve general image quality including prompt alignment and aesthetics. Popular DPO methods propagate preference labels from clean image pairs to all the intermediate steps along the two generation trajectories. However, preference labels provided in existing datasets are blended with layout and aesthetic opinions, which would disagree with aesthetic preference. Even if aesthetic labels were provided (at substantial cost), it would be hard for the two-trajectory methods to capture nuanced visual differences at different steps. To improve aesthetics economically, this paper uses existing generic preference data and introduces step-by-step preference optimization (SPO) that discards the propagation strategy and allows fine-grained image details to be assessed. Specifically, at each denoising step, we 1) sample a pool of candidates by denoising from a shared noise latent, 2) use a step-aware preference model to find a suitable win-lose pair to supervise the diffusion model, and 3) randomly select one from the pool to initialize the next denoising step. This strategy ensures that diffusion models focus on the subtle, fine-grained visual differences instead of layout aspect. We find that aesthetics can be significantly enhanced by accumulating these improved minor differences. When fine-tuning Stable Diffusion v1.5 and SDXL, SPO yields significant improvements in aesthetics compared with existing DPO methods while not sacrificing image-text alignment compared with vanilla models. Moreover, SPO converges much faster than DPO methods due to the use of more correct preference labels provided by the step-aware preference model. Code and models are available at \href{https://github.com/RockeyCoss/SPO}{https://github.com/RockeyCoss/SPO}.
\end{abstract}

\section{Introduction}
\label{sec:intro}

\begin{figure*}[t]
\centering
\includegraphics[width=1\textwidth]{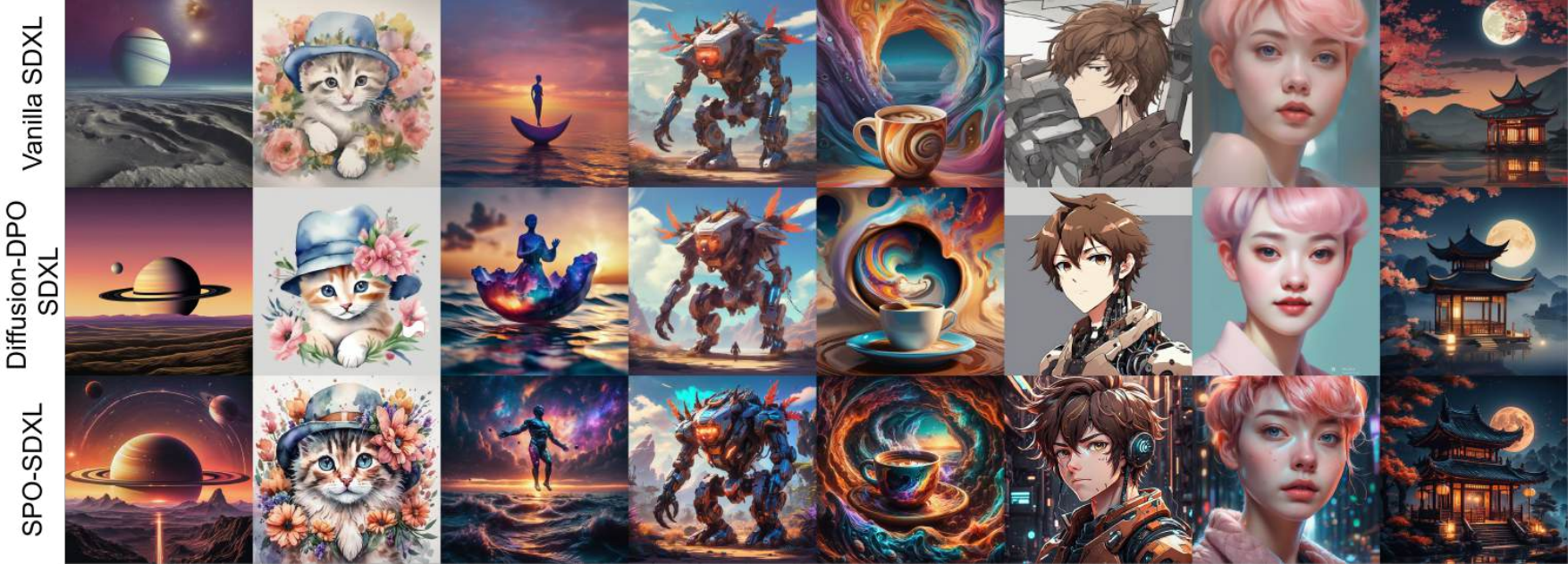}
\caption{{Qualitative comparison between {Vanilla SDXL, Diffusion-DPO-SDXL and SPO-SDXL. SPO-SDXL exhibits very strong image aesthetics including more visual details and appealing styles.} Prompts are provided in the supplementary material.}}
\label{fig:qualitative_comp}
\end{figure*}

This paper aims to improve the ability of diffusion models in generating visually appealing images based on human preference data. That is, given a pool of image pairs and human preference within each pair, we fine-tune diffusion models so that they are more likely to generate images consistent with human aesthetic preference\footnote{We fine-tune models using crowd-sourced preference data. We do not discuss individual, cultural or political impact here.}. 

Direct preference optimization (DPO) is shown to be effective for aligning diffusion models with human preferences in general aspects such as image-text alignment \cite{diffusion_dpo}. Given a pair of images generated from the same prompt, 
DPO methods encourage predictions to align with the preferred image while discouraging resemblance to the dispreferred one. A few existing studies use human preferences at timestep 0 with clean images $\boldsymbol{x}^w_{0}$ and $\boldsymbol{x}^l_{0}$, where $w$ and $l$ are win and lose preference labels, respectively \citep{rafailov2024direct, d3po, diffusion_dpo}. The labels are directly propagated to intermediate samples along the two trajectories, assuming all intermediate samples along the win trajectory are also preferred. 

However, for two problems existing systems are not best effective for aesthetic alignment. \underline{First}, generic preferences provided in public datasets like Pick-a-Pic \cite{kirstain2024pick} are not aesthetics specific, and they often disagree. In Fig. \ref{fig:trajc_problem_vis}, (a) is generally preferred because of the correctly generated dog, cat, and their spatial arrangement, but in terms of aesthetics, (a) should be dispreferred. More examples are provided in the supplementary material. This introduces noisy supervision signals and compromises the model improvement towards better aesthetics. Technically, this problem can be fixed by manually annotating an aesthetics-only preference dataset, but it is very expensive, because of layout influence and complexity of aesthetics. 
\underline{Second}, in two-trajectory methods, the images within each pair at each denoising step look very different, as shown in Fig. \ref{fig:trajc_problem_vis}. Even if accurate aesthetic preference was provided at each step (at great cost), it would still be very non-trivial to learn, because  the large layout discrepancy would dominate over aesthetic nuances. 

To better align diffusion models with aesthetic preference, we introduce step-by-step preference optimization (SPO). 
SPO is new in that it completely discards the current preference propagation strategy, pushing for evaluating image details. 
Specifically, at each step beginning with a noisy image $\boldsymbol{x}_{t}$, we generate a pool of $\boldsymbol{x}_{t-1}$ samples. We evaluate their quality using {a step-aware preference model (SPM)} and assign win/lose labels to the pair showing the largest quality difference. We then randomly select an image from the pool to initialize timestep $t-1$. Because the win-lose pair 1) comes from the same image and 2) is generated in one or very few steps, the two samples would only exhibit small differences in details. SPM allows us to capture such detailed differences and guide the diffusion model to generate more visually pleasing images.

We use SPO to fine-tune Stable Diffusion v1.5 (SD-1.5)~\citep{rombach2022high} and SDXL~\citep{podell2023sdxl}. Although we use a training set with generic human perferences~\cite{kirstain2024pick}, we demonstrate significant improvement in aesthetics compared with those fine-tuned by popular DPO methods. 
Moreover, SPO converges much faster than Diffusion-DPO. This is because the step-by-step design makes it easier to focus on fine-grained visual details, and SPM produces more accurate preference labels. 
We summarize key points of this paper below.
\begin{itemize}[leftmargin=*]
    \item[•] We aim to improve image aesthetics of diffusion models.
    \item[•] We point out that generic human preferences are often different from pure aesthetic preference and that obtaining aesthetic-only preference data is very expensive.  
    \item[•] We design SPO, where we determine win-lose pairs at each step in an online manner. We make sure win-lose pairs come from the same noisy sample, so after one or very few steps of denoising their differences are small and in fine-grained details, and can be captured by a preference model trained with generic preference data. 
    \item[•] When fine-tuning SD-1.5 and SDXL, SPO is more effective in enhancing image aesthetics compared with DPO methods, and converges faster than Diffusion-DPO \citep{diffusion_dpo}. 
\end{itemize}

\section{Related Work}
Recently, inspired by post-training methods that improve LLMs, \textit{e.g.}, reinforcement learning from human feedback (RLHF)~\citep{ouyang2022training,zhang2024large}, various post-training methods are proposed to align pre-trained diffusion models with human preferences.
For example, 
\citet{chen2023enhancing} leverage 
the PPO loss~\citep{schulman2017proximal} to fine-tune the text encoder of diffusion models. AligningT2I~\citep{lee2023aligning} develops a reward model to evaluate the quality of generated images and fine-tunes the diffusion model using image-text pairs, weighted by assessment of the reward model. DPOK~\citep{fan2024reinforcement} and DDPO~\citep{black2023training} use policy gradient to fine-tune diffusion models, aiming at maximizing reward signals. Furthermore, ReFL~\citep{xu2024imagereward}, DRaFT~\citep{clark2023directly}, and AlignProp~\citep{prabhudesai2023aligning} directly propagate the gradients through differentiable reward models to fine-tune the denoising steps.
Recent methods are inspired by direct preference optimization (DPO)~\citep{rafailov2024direct}, which eliminates the need for explicit reward models when post-training LLMs. Diffusion-DPO~\citep{diffusion_dpo} fine-tunes diffusion models on the Pick-a-Pic~\citep{kirstain2024pick} dataset containing image preference pairs.
D3PO~\citep{d3po} generates pairs of images from the same prompt and uses a preference model to identify preferred and dispreferred images. DenseReward~\citep{yang2024dense} improves the DPO scheme with a temporal discounting approach to emphasize initial denoising steps. 
{These methods optimize the trajectory-level preference, where the accumulated image differences are too large to allow the network to focus on aesthetics subtleties. In comparison, SPO by its step-by-step mechanism can focus on nuanced visual differences in just a single or few steps.}

\section{DPO Revisit and Diagnosis}

\textbf{General formulation.} Given a generation model $\pi_\theta(\cdot)$ and a condition $c$, the probability of generating output $o$ is $\pi_\theta(o \mid c)$. We use $\pi_\theta(\cdot)$ to generate a set of output pairs $\mathcal{S}$, where each pair comes from the same condition $c$. Human or preference model is employed to label the preference order of the output pairs as $(o^w, o^l, c)$, where $o^w$ is the preferred output and $o^l$ is dispreferred. According to~\citet{rafailov2024direct}, the DPO loss used to fine-tune $\pi_\theta$ is defined as:
\begin{equation}
\label{eq:general_dpo_loss}
\scalebox{0.89}{$
     \mathcal{L}_{\mathcal{DPO}} =  - 
    \mathbb{E}_{(o^w, o^l, c) \sim \mathcal{S}} \left[
    \log\sigma\left(\beta \log \frac{\pi_{\theta}\left(o^w \mid c\right)}{\pi_{\textrm{ref}}\left(o^w \mid c\right)}-\beta \log \frac{\pi_{\theta}\left(o^l \mid c\right)}{\pi_{\textrm{ref}}\left(o^l \mid c\right)}\right)\right].
$}
\end{equation}
$\pi_{\textrm{ref}}(\cdot)$ and $\sigma(\cdot)$ refer to the reference model and the sigmoid function, respectively. $\beta$ is the strength of regularization.

\textbf{DPO for diffusion model post-training.} We denote the text-to-image diffusion model with parameters $\theta$ as $p_\theta$ and text prompt as $c$. The denoising process generates intermediate states $\{\boldsymbol{x}_T, \boldsymbol{x}_{T-1}, ..., \boldsymbol{x}_1, \boldsymbol{x}_0\}$. Existing works including Diffusion-DPO \citep{diffusion_dpo} and D3PO \citep{d3po} measure preference based on the final image $\boldsymbol{x}_0$ and assign the preference for $\boldsymbol{x}_0$ directly to the entire generation trajectory, or all the intermediate states. Let $\mathcal{T}_w$ and $\mathcal{T}_l$ denote the denoising trajectories which generate the preferred and dispreferred images, respectively. Using the Markov property of diffusion models and Jensen’s inequality, they reformulate the general DPO loss in Eq. \ref{eq:general_dpo_loss} into the following step-wise form:
\begin{equation}
\label{eq:dpo_in_diff}
\scalebox{0.7}{$
\mathcal{L}_\mathcal{DPO-D} =  - 
\mathbb{E}_{\substack{(\boldsymbol{x}^w_{t-1}, \boldsymbol{x}^w_{t}) \sim \mathcal{T}_w \\ (\boldsymbol{x}^l_{t-1}, \boldsymbol{x}^l_{t}) \sim \mathcal{T}_l}}  \left[
    \log\sigma\left(\beta \log \frac{p_{\theta}\left(\boldsymbol{x}^w_{t-1} \mid \boldsymbol{x}^w_{t},c\right)}{p_{\textrm{ref}}\left(\boldsymbol{x}^w_{t-1} \mid \boldsymbol{x}^w_{t},c\right)} \right.\right. 
    \left.\left. -\beta \log \frac{p_{\theta}\left(\boldsymbol{x}^l_{t-1} \mid \boldsymbol{x}^l_{t},c\right)}{p_{\textrm{ref}}\left(\boldsymbol{x}^l_{t-1} \mid \boldsymbol{x}^l_{t},c\right)}\right)\right],
$}
\end{equation}
where $\mathcal{L}_\mathcal{DPO-D}$ encourages denoising steps to progress towards the preferred image and away from dispreferred one.

\begin{figure*}
\centering
\includegraphics[width=1.0\textwidth]{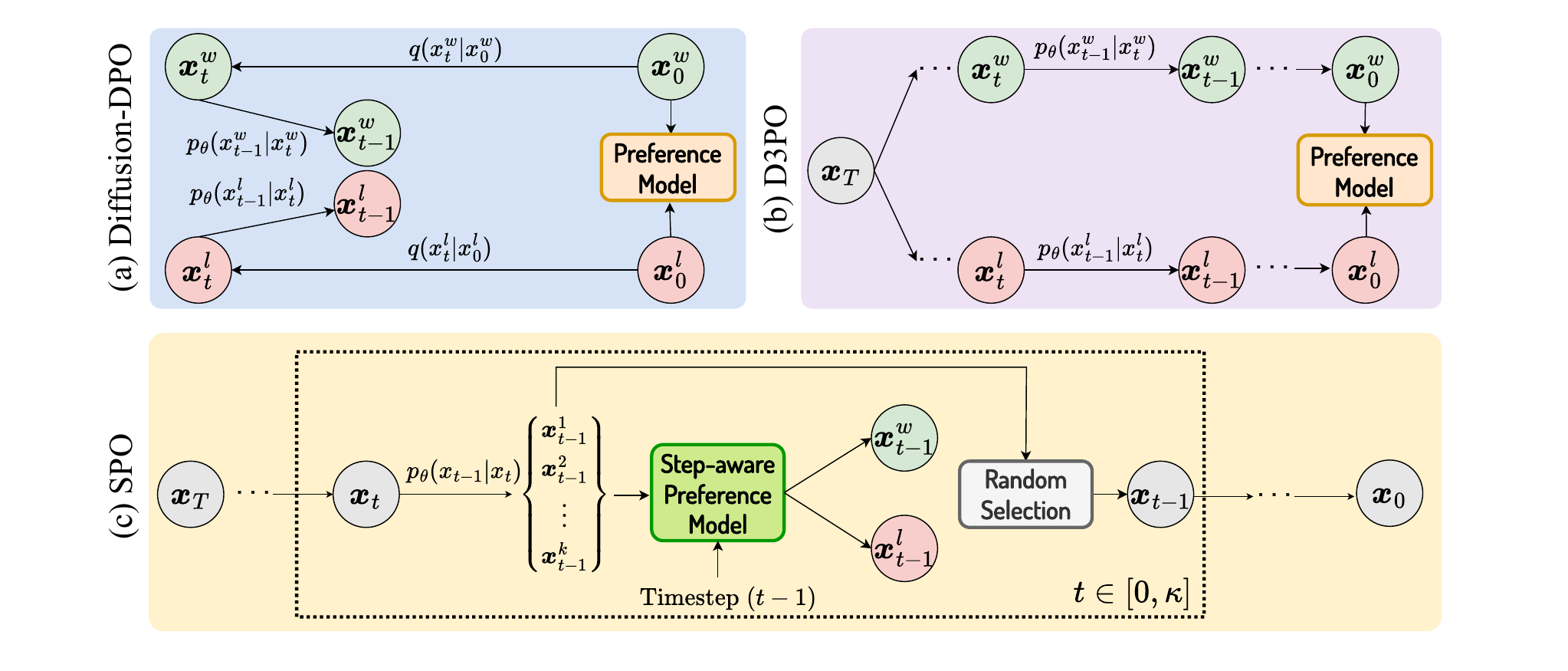}

\caption{{Comparing frameworks of SPO, Diffusion-DPO, and D3PO approaches. SPO does not adopt direct preference propagation as other DPO methods do. In SPO, a pool of samples are generated at each step, from which a proper win/lose pair is selected and used to fine-tune the diffusion model. Then, a single sample is randomly selected to initialize the next iteration.}}
\label{fig:comparision_dpo_methods}
\end{figure*}

\textbf{Diagnosis on aesthetic alignment}. Aligning diffusion models with human aesthetic preference is very challenging. From architecture, win-lose pairs in the two-trajectory methods usually differ significantly (primarily in layout), rendering it hard to focus on image detail comparisons. From data, existing aesthetic scoring datasets~\cite{murray2012ava, laionaes} do not provide paired image data coming from the same prompt. Collecting a dedicated aesthetic preference dataset would be more costly compared with tasks like classification or prompt alignment preference. So a cost-efficient choice is existing generic preference datasets where annotators were asked to score images based on holistic opinions like prompt alignment and aesthetics. 
However as pointed out in Section \ref{sec:intro}, generic preferences may be contradictory to aesthetic preference. If we directly propagate them to all diffusion steps like Diffusion-DPO and D3PO do (see Fig. \ref{fig:comparision_dpo_methods}), noisy preferences will compromise fine-tuning.

\section{Proposed Approach}

\subsection{Framework Overview}\label{sec:framework}
To align diffusion models with aesthetic preference while still using generic preferences, we propose step-by-step preference optimization (SPO), an online reinforcement learning method. Fig. \ref{fig:comparision_dpo_methods} $(c)$ depicts its workflow.
Given an intermediate $\boldsymbol{x}_{t}$, at timestep $t$, we sample a pool of denoised samples  $\{\boldsymbol{x}_{t-1}^1, \boldsymbol{x}_{t-1}^2, \cdots, \boldsymbol{x}_{t-1}^k\}$. 
We apply a step-aware preference model (Section \ref{sec:step_aware_preference_model}) to compare the quality of these candidate samples, and select the highest-quality sample and the lowest-quality sample as the win-sample and the lose-sample, respectively. 
SPO ensures the win-lose pair comes from the same $\boldsymbol{x}_{t}$ and thus have small differences in image details to reflect aesthetics. 
We then randomly select a sample from the candidate pool (Section \ref{sec:step_wise_resampler}), which is used to initialize the next iteration. SPO is optimized by a revised DPO loss function (Section \ref{sec:step_aware_objective}) and can be extended to the SDXL model (Section \ref{sec:multistep_spo}).

\subsection{Step-Aware Preference Model (SPM)}\label{sec:step_aware_preference_model}

\textbf{Overall use}. SPM predicts preference order among the $k$ sampled denoised samples $\{\boldsymbol{x}_{t-1}^1, \boldsymbol{x}_{t-1}^2, \cdots, \boldsymbol{x}_{t-1}^k\}$. Thus, SPM takes timestep $t-1$, intermediate sample $\boldsymbol{x}_{t-1}$, and prompt $c$ as input, and outputs a quality score. SPM is different from existing preference models. The latter uses clean images $\boldsymbol{x}_{0}$ and prompt $c$ as input without time condition, because they are designed for assessing clean images. Following \citep{kirstain2024pick}, we construct SPM based on CLIP \citep{radford2021learning}.

For \textbf{SPM training}, we initialize the model with PickScore~\citep{kirstain2024pick} and fine-tune it following \citep{dhariwal2021diffusion}, making the model useful for noisy images. Specifically, we add the same amount of noise to a pair of clean images, assuming that their preference order can be kept. During training, for each pair of images and their human-labeled preference, we randomly sample a timestep $t$ and add the same noise to both images to obtain $\boldsymbol{x}^w_{t}$ and $\boldsymbol{x}^l_{t}$. 
Then we feed the noisy intermediate pair $\{\boldsymbol{x}^w_{t}, \boldsymbol{x}^l_{t}\}$ and timestep $t$ to SPM and train the model to correctly predict the preference, using loss function \textcolor{black}{$\mathcal{L}_{\text{pref}} = \left( \log 1 - \log \hat{p}_w\right)$}, where $\hat{p}_w$ is the probability of the win image being the preferred one, following~\cite{kirstain2024pick}. $\hat{p}_w$ is computed using the following equation:
\begin{equation}
\scalebox{0.90}{$
\hat{p}_w = \frac{ \exp{(\tau \cdot f_{\textrm{CLIP-V}}( \boldsymbol{x}^w_{t}, t) \cdot f_{\textrm{CLIP-T}}(c))}}{\exp{(\tau \cdot f_{\textrm{CLIP-V}}( \boldsymbol{x}^w_{t}, t) \cdot f_{\textrm{CLIP-T}}(c))} + \exp{(\tau \cdot f_{\textrm{CLIP-V}}( \boldsymbol{x}^l_{t}, t) \cdot f_{\textrm{CLIP-T}}(c))}},$}
\end{equation}
where $c$ represents the text prompt, $\tau\in \mathcal{R}$ is a temperature. $f_\textrm{CLIP-V}(\cdot)$ and $f_\textrm{CLIP-T}(\cdot)$ are the vision and text encoders of CLIP, respectively. 
To allow for timestep-conditional preference prediction, we modify the CLIP vision encoder using time-conditional adaptive layernorm~\citep{peebles2023scalable}. To alleviate the domain gap between the noised image at the $t$-th timestep and images used to train the PickScore model, we estimate $\boldsymbol{\hat{x}}_{0}$ from the noisy sample directly, following DDIM~\cite{song2020denoising}.

\begin{figure*}
\centering
\includegraphics[width=1.0\textwidth]{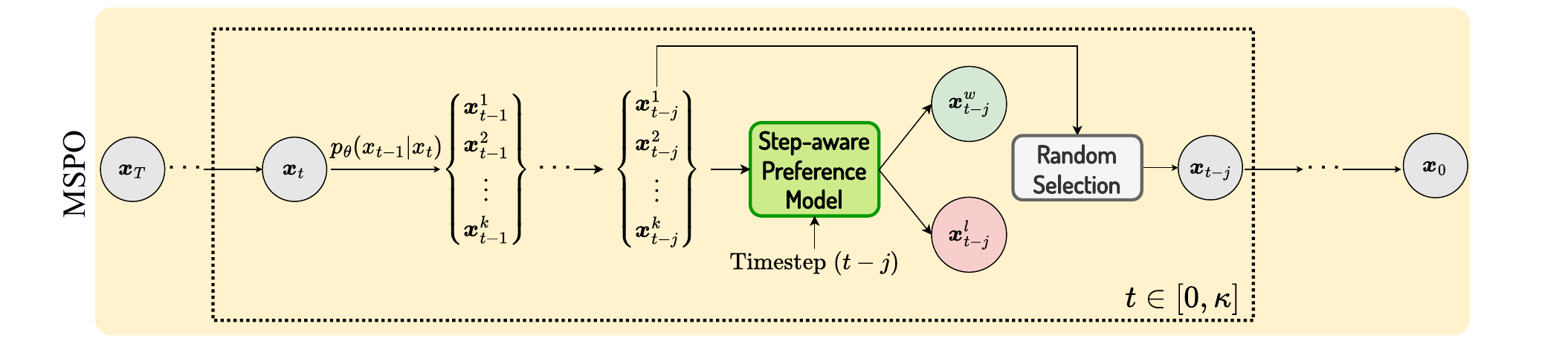}
\caption{Framework of multi-step-by-step preference optimization (MSPO). From $\boldsymbol{x}_{t}$, we first sample $k$ latents $\{\boldsymbol{x}_{t-1}^1, \boldsymbol{x}_{t-1}^2, \cdots, \boldsymbol{x}_{t-1}^k\}$ as SPO does. For each latent, we perform multiple (\textit{i.e.}, $j$) denoising steps to obtain $\{\boldsymbol{x}_{t-j}^1, \boldsymbol{x}_{t-j}^2, \cdots, \boldsymbol{x}_{t-j}^k\}$, from which a win-lose pair is selected by SPM. 
Then we apply random selection and iterations in the same manner as SPO.}
\label{fig:multistep_spo}
\end{figure*}

\subsection{Random Selection of $\boldsymbol{x}_{t}$ to Start Next Step}\label{sec:step_wise_resampler}
In SPO, we randomly sample $\boldsymbol{x}_{t}$ from the candidate pool $\{\boldsymbol{x}_{t}^1, \boldsymbol{x}_{t}^2, \cdots, \boldsymbol{x}_{t}^k\}$, which is then used to start the next denoising step. This ensures every win-lose pair comes from the same latent.
Since $\boldsymbol{x}_{t}$ is very noisy when $t$ is large, we use the standard diffusion sampling process when $t$ is greater than the threshold $\kappa$. Only when $t \leq \kappa$ do we sample candidate pools and apply random selection. 
This random selection is shown in Fig. \ref{fig:comparision_dpo_methods} $(c)$: selection of $\boldsymbol{x}_{t-1}$ is depicted and is consistent with the above discussion.

\subsection{Objective Function of SPO}\label{sec:step_aware_objective}
At the $t$-th denoising timestep, we sample a pool of candidates $\{\boldsymbol{x}_{t-1}^1, \boldsymbol{x}_{t-1}^2, \cdots, \boldsymbol{x}_{t-1}^k\}$ and use the step-aware preference model to construct a preference pair $(\boldsymbol{x}^w_{t-1}, \boldsymbol{x}^l_{t-1})$, where $\boldsymbol{x}^w_{t-1}$ and $\boldsymbol{x}^l_{t-1}$ are the most and least preferred in the pool. Using various prompts, we can obtain many preference pairs at $t$-th timestep. Using the general form of DPO loss in Eq. \ref{eq:general_dpo_loss}, the {SPO} objective at the $t$-th timestep is:
\begin{equation}
\label{eq:step_dpo_loss}
\begin{aligned}[t]
& \scalebox{0.94}{$\mathcal{L}_t(\theta) =  - 
\mathbb{E}_{c \sim p(c), \boldsymbol{x}^w_{t-1}, \boldsymbol{x}^l_{t-1} \sim p_{\theta}\left(\boldsymbol{x}_{t-1} \mid c, t, \boldsymbol{x}_{t}\right)}$} \\
    & \scalebox{0.94}{$\left[
    \log\sigma\left(\beta \log \frac{p_{\theta}\left(\boldsymbol{x}^w_{t-1} \mid c, t, \boldsymbol{x}_{t}\right)}{p_{\textrm{ref}}\left(\boldsymbol{x}^w_{t-1} \mid c, t, \boldsymbol{x}_{t}\right)} \right.\right. \left.\left. -\beta \log \frac{p_{\theta}\left(\boldsymbol{x}^l_{t-1} \mid c, t, \boldsymbol{x}_{t}\right)}{p_{\textrm{ref}}\left(\boldsymbol{x}^l_{t-1} \mid c, t, \boldsymbol{x}_{t}\right)}\right)\right],$}
\end{aligned}
\end{equation}
where $c$ is the prompt and $p(c)$ is the distribution of prompts. 
By combining the {SPO} objectives across all $T$ timesteps, we obtain the final {SPO} objective:
\begin{multline}
    \scalebox{0.93}{$\mathcal{L}(\theta)= - 
    \mathbb{E}_{t \sim \mathcal{U}[1, T-\kappa], c \sim p(c), x_T \sim \mathcal{N}(\mathbf{0}, \mathbf{I}), \boldsymbol{x}^w_{t-1}, \boldsymbol{x}^l_{t-1} \sim p_{\theta}\left(\boldsymbol{x}_{t-1} \mid c, t, \boldsymbol{x}_{t}\right)}$} \\
    \scalebox{0.95}{$
    \left[
    \log\sigma\left(\beta \log \frac{p_{\theta}\left(\boldsymbol{x}^w_{t-1} \mid c, t, \boldsymbol{x}_{t}\right)}{p_{\textrm{ref}}\left(\boldsymbol{x}^w_{t-1} \mid c, t, \boldsymbol{x}_{t}\right)}-\beta \log \frac{p_{\theta}\left(\boldsymbol{x}^l_{t-1} \mid c, t, \boldsymbol{x}_{t}\right)}{p_{\textrm{ref}}\left(\boldsymbol{x}^l_{t-1} \mid c, t, \boldsymbol{x}_{t}\right)}\right)\right],$}
    \label{eq:final_loss_diffusion}
\end{multline}
where $\mathcal{U}$ and $\mathcal{N}$ are the uniform distribution and Gaussian distribution, respectively.

\subsection{Extension to Multi-step Preference Optimization for SDXL}\label{sec:multistep_spo}

For stronger models like SDXL, we observe that the difference between $\boldsymbol{x}_t$ and $\boldsymbol{x}_{t-1}$ is \textit{too small}, so is the difference between the selected candidates $\boldsymbol{x}^w_{t-1}$ and $\boldsymbol{x}^l_{t-1}$. While a relatively small difference allows SPM to focus on image details, that difference is too small would create ambiguities and confuse fine-tuning. To address this issue, we expand step-by-step preference optimization to multi-step preference optimization (MSPO), which uses multiple denoising steps to increase the diversity of the candidate set (see Fig.~\ref{fig:multistep_spo}). This simple extension allows us to select more different samples and more clear preference signals. 

\subsection{Discussions and Insights}\label{sec:discussion}

\begin{figure}[t]
\begin{minipage}[t]{1\linewidth}
\centering
\includegraphics[width=1\textwidth]{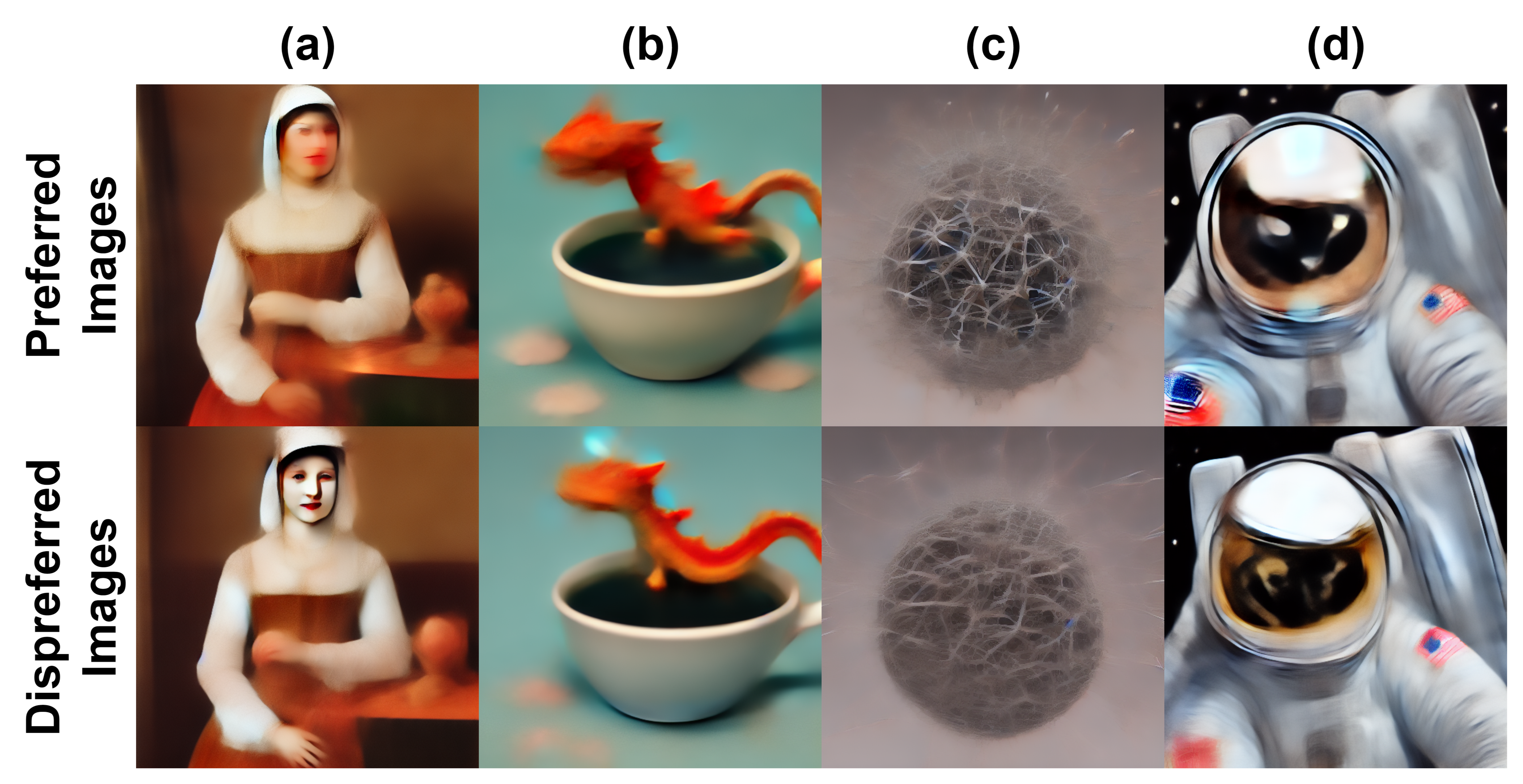}
\caption{Win-lose pairs identified by the SPM during fine-tuning. Top: Preferred images. Bottom: Dispreferred images. In SPO, these pairs look similar so image details can be focused on. From these pairs, SPM favors images with fewer artifacts in (a) and (b) and more refined details in (c) and (d). These images appear blurry because for visualization purpose we directly predict the clean image from the intermediate noisy latents.}
\label{fig:training_preference_pairs}
\end{minipage}
\end{figure}

\textbf{DPO for diffusion model vs. language model.} Diffusion models involve many intermediate steps, each producing a latent feature/image. In contrast, language models typically predict the final result at each position with a single step without iterative refinement. This distinction necessitates specific DPO design in diffusion models.

\textbf{SPO is an implicit aesthetic optimizer and distillator.} We do not use a dedicated aesthetic preference dataset. Aesthetic optimization is done implicitly through SPM, which is trained to perceive image quality from generic opinions. When the two images to be compared are relatively similar, the output from SPM thus mainly describes image details instead of significant layout differences. Sample win-lose pairs selected by the SPM are shown in Fig. \ref{fig:training_preference_pairs}.  In this way, we are able to distill aesthetic details from generic data.

\textbf{How often does generic preferences and aesthetic preference disagree?} It is hard to quantify because it is non-trivial to annotate a validation set. {Fig. \ref{fig:trajc_problem_vis} and supplementary material depict some disagreement scenarios based on careful manual inspection.} In fact, because of the nuanced nature of aesthetics, an image may be superior in certain aesthetic aspects while its paired image is better in other aspects. So there are probably more disagreement scenarios than we can think of now.  

\textbf{Limitations.} First, SPO is not applicable to recent flow matching models, such as {SD3~\cite{esser2024scaling} and Flux~\cite{flux} because SPO requires the intermediate steps to be stochastic which flow matching models do not satisfy. 
Second, SPO is specifically designed to improve aesthetics and offers limited help for improving image-text alignment. Third, we also limit ourselves to learning from crowd-sourced data without tapping into the subjective, political, and historical aspects of aesthetics. It is interesting to study these problems in future.}

\section{Experiments}

\subsection{Experimental Setup}\label{sec:setup}
\textbf{Datasets.} {For SPO, we train the step-aware preference model (SPM) on the Pick-a-Pic V1 dataset. This dataset has over 580k labeled image preference pairs, each generated by the same text prompt with various diffusion models~\cite{stable-diffusion-v1-5, stable-diffusion-v2-1, podell2023sdxl}. Human annotators were asked to rate the \textit{general} quality of each image, forming win-lose pairs. When fine-tuning diffusion models with SPO, DDPO or D3PO, we use a subset of 4k prompts (without images) randomly selected from Pick-a-Pic V1, where win-lose pairs are generated online. Note that these three methods do not use images or preference labels but use text prompts only.}
For DDPO and D3PO, PickScore trained on Pick-a-Pic V1 is used as their reward model to provide guidance. For Diffusion-DPO and MAPO~\cite{she2024mapo}, we use their online-available checkpoints for evaluation, which were trained on Pick-a-Pic V2 dataset with over 800k image preference pairs. So overall, SPO and the competitive models are trained on similar datasets and can be fairly compared.

If not specified, we report quantitative results based on the $500$ validation prompts, \textit{i.e.}, \texttt{validation\_unique} split of Pick-a-Pic~\cite{kirstain2024pick}, which is adopted in \cite{diffusion_dpo}. {We also use GenEval~\cite{ghosh2024geneval} to evaluate image-text alignment, including rendering of single and two objects, counting, colors, position and attribute binding. There are 553 test prompts.}

\textbf{Evaluation Protocol.} {This paper evaluates image quality with four automatic metrics. We use PickScore \cite{kirstain2024pick}, HPSV2 \cite{wu2023human} and ImageReward \cite{xu2024imagereward} for prompt-aware human preference estimation. These models are trained on human preference datasets and learn to replicate human decisions about which images are more favorable. We use Aesthetic score \cite{laionaes} to evaluate visual appeal. This score is prompt agnostic and employs a linear estimator on top of the vision encoder of CLIP~\cite{radford2021learning} to predict the aesthetic quality of images. Note that PickScore, HPSV2, and ImageReward all assess aesthetics to some extent. Apart from these automatic metrics, we also use human studies. We invite 10 people to assess 300 pairs of images generated by  two methods of interest. Their preferences are summarized into winning percentage from 0 to 100\%. The text prompts are randomly selected from PartiPrompts (100 prompts) \cite{yu2022scaling} and HPSV2 benchmark (200 prompts) \cite{wu2023human}.}

\setlength{\tabcolsep}{1.68mm}
\begin{table}[t]
\centering 
\small
\caption{
Method comparison on SDXL. SPO overall yields the \textbf{best} fine-tuning performance especially in aesthetics. {Note PickScore, HPSV2, and ImageReward partially assess aesthetics.}}
\label{tab:prompt_alignment_matrix_sdxl}  
\begin{tabular}{l|cccc}  
\hline
    Method     &  PickScore & HPSV2 & ImageReward & Aesthetic\\
    \hline 
    SDXL    & 21.95 & 26.95 & 0.5380 & 5.950  \\
    Diff.-DPO  & 22.64 & 29.31 & 0.9436 & 6.015  \\
    MAPO & 22.11 & 28.22 & 0.7165 & 6.096 \\
    SPO      & \textbf{23.06} & \textbf{31.80} & \textbf{1.0803} & \textbf{6.364} \\
    \hline
\end{tabular}
\end{table}

\setlength{\tabcolsep}{1.68mm}
\begin{table}[t]
\centering 
\small
\caption{Comparing different methods on SD-1.5.}
\label{tab:auto_eval_matrix_sd15}  
{
\begin{tabular}{l|cccc}  
\hline
    Method     &  PickScore & HPSV2 & ImageReward & Aesthetic \\
    \hline  
    SD-1.5    & 20.53 & 23.79 & -0.1628 & 5.365 \\
    DDPO      & 21.06 & 24.91 & 0.0817 & 5.591 \\
    D3PO      & 20.76 & 23.97 & -0.1235 & 5.527   \\
    Diff.-DPO  & 20.98 & 25.05 & 0.1115 & 5.505  \\
    SPO     & \textbf{21.43} & \textbf{26.45} & \textbf{0.1712} & \textbf{5.887} \\
    \hline

\end{tabular}  
}
\end{table}

\setlength{\tabcolsep}{0.8mm} 
\begin{table}[t]
\centering
\footnotesize 
\caption{Method comparison on GenEval based on SDXL. $\dagger$: results are reproduced with a classifier-free guidance scale of 5.0 and 50 inference steps. GenEval evaluates image-text alignment.}
\label{tab:geneval_results}
\begin{tabular}{l|ccccccc}
\toprule
           & Single & Two     &         &        &          & Attribute & \\
    Method & Object & Object & Counting & Colors & Position & Binding   & Overall \\
\hline
    SDXL$\dagger$       & 97.50 & 71.21 & 39.06 & 84.04 & 11.00 & 17.75 & 53.43 \\
    Diff.-DPO         & \textbf{99.06} & \textbf{80.81} & \textbf{46.56} & \textbf{88.30} & \textbf{13.25} & \textbf{29.50} & \textbf{59.58} \\
    SPO               & 97.81 & 73.48 & 41.25 & 85.64 & 13.00 & 20.00 & 55.20 \\
\bottomrule
\end{tabular}
\end{table}

\textbf{Implementation details.} 
We apply DDIM~\cite{song2020denoising} with $\eta=1.0$ and 20 timesteps as the sampler and use classifier free guidance~\cite{ho2022classifier} with scale 5.0 for sampling during online training. We use LoRA~\cite{hu2021lora} for both SD-1.5 and SDXL, fine-tuning the models for $10$ epochs. The LoRA rank is 4 and 64 for SD-1.5 and SDXL, respectively.
We set the strength of regularization $\beta=10$. For SD-1.5, we set the batch size as $40$ and  learning rate as $6e^{-5}$. For SDXL, we set the batch size as $8$, gradient accumulation as $2$, and learning rate as $1e^{-5}$. Since very noisy images are difficult to compare, we do not use SPM to very early stages. That is, 
we only compute the preference of $\boldsymbol{x}_{t}$ when $t \leq \kappa$ and consider all $\boldsymbol{x}_{t}$ with $t > \kappa$ as tied. We empirically set $\kappa$ as $750$ and will evaluate $\kappa$ in Section \ref{sec:ablation}. When fine-tuning SDXL with the MSPO {(Section \ref{sec:multistep_spo})}, 
we set the number of inner steps to $4$. We do not apply SDXL refiner~\cite{podell2023sdxl} to ensure fair comparison. {{For SPM training, we adopt learning rates of 3e-6 and 1e-6 for SD-1.5 and SDXL, respectively.}} We use DDIM scheduler with classifier free guidance scale of 5 and 20 steps to perform inference on validation prompts.

\begin{figure*}
\centering
\includegraphics[width=1\textwidth]{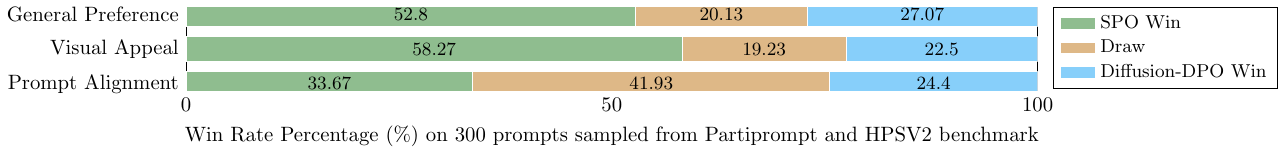}
\includegraphics[width=1\textwidth]{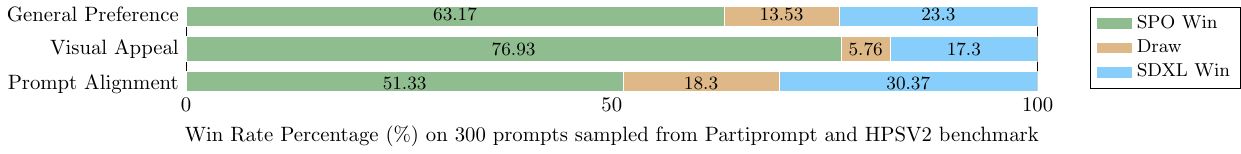}
\caption{{User study results comparing SPO with Diffusion-DPO and Vanilla SDXL. We sampled 100 and 200 prompts for evaluation from Partiprompts~\cite{yu2022scaling} and HPSV2 benchmark~\cite{wu2023human}, respectively. SPO yields clear improvement in visual appeal.}} 
\label{fig:win_rate}
\end{figure*}

\begin{table*}[!ht]
\centering
\begin{minipage}[t]{0.33\linewidth}
\setlength{\tabcolsep}{1.68mm}
\centering
\footnotesize
\caption{{Comparing random sampling with other sampling strategies.}} 
\label{tab:reinitialization}  
{
\begin{tabular}{l|cccc}  
\hline
Initial.  &  P-S & HPSV2 & I-R & AE\\
\hline  
$\boldsymbol{x}^{w}_{t-1}$   &  17.87 & 11.31 & -2.2692 & 3.963\\
$\boldsymbol{x}^{l}_{t-1}$   &  19.36 & 18.63 & -1.3743 & 5.338 \\
random    & \textbf{21.43} & \textbf{26.45} & \textbf{0.1712} & \textbf{5.887} \\
\hline
\end{tabular}
}
\end{minipage}
\begin{minipage}[t]{0.33\linewidth}
\setlength{\tabcolsep}{1.08mm}
\centering
\footnotesize
\caption{{Comparing SPM with variants: no time condition or the PickScore model \cite{kirstain2024pick}.}} 
\label{tab:step_aware_preference_model}
{
\begin{tabular}{l|cccc}
\hline
Prefer. model     &  P-S & HPSV2 & I-R & AE\\
    \hline  
    {SPM}  & \textbf{21.43} & \textbf{26.45} & \textbf{0.1712} & \textbf{5.887} \\
    {w/o step con.}   &  21.19 & 25.84 &  0.1365 & 5.678 \\
    {PickScore} &20.28 &23.09 &-0.2982 &5.410\\
    \hline
\end{tabular}
}
\end{minipage}
\begin{minipage}[t]{0.33\linewidth}
\setlength{\tabcolsep}{1.08mm}
\centering
\footnotesize
\caption{{Impact of number of sampled images $k$ at each step. \textcolor{black}{We use $k=4$.}}}
\label{tab:num_of_sample_per_step_ablation}
{
\begin{tabular}{l|cccc}  
\hline
  \#{}samples $k$ &  P-S & HPSV2 & I-R & AE\\ 
    \hline  
    2 & 21.37 & 26.56 & 0.3235 & 5.714 \\
    4 & \textbf{21.43} & 26.45 & 0.1712 & \textbf{5.887} \\
    8 &  21.19 & \textbf{27.62} & \textbf{0.4199} & 5.605\\
    \hline
\end{tabular}
}
\end{minipage}
\begin{minipage}[t]{0.33\linewidth}
\setlength{\tabcolsep}{1.08mm}
\centering
\footnotesize
\caption{Impact of \#{}inner steps $j$ when fine-tuning SDXL with MSPO. We set $j=4$.}
\label{tab:num_inner_step}
{
\begin{tabular}{c|cccc}  
\hline
  \#{}inner steps $j$  & P-S & HPSV2 & I-R & AE\\  
    \hline  
    1   &  22.85 & 31.37 & 1.0071 & 6.359 \\
    2  &  22.84 & 31.17 & 1.0118 & 6.268 \\
    3   &  22.94 & 31.55 & \textbf{1.0847} & 6.380 \\
    4   &  \textbf{23.06} & \textbf{31.80} & 1.0803 & 6.364 \\
    5   &  23.03 & 31.23 & 0.9656 & \textbf{6.423} \\
    6   &  22.95 & 30.57 & 0.9770 & 6.390\\
    \hline
\end{tabular}  
}
\end{minipage}
\begin{minipage}[t]{0.33\linewidth}
\setlength{\tabcolsep}{0.85mm}
\centering
\footnotesize
\caption{Impact of timestep range.}
\label{tab:timestep_range}
{
\begin{tabular}{c|cccc} 
\hline
  Timestep Range  &  P-S & HPSV2 & I-R & AE\\  
    \hline  
    \texttt{[0-250]}   &  20.61 & 23.34 & -0.1823 & 5.413 \\
    \texttt{[0-500]}   &  20.69 & 25.67 & 0.0810 & 5.399 \\
    \texttt{[0-750]}   &  \textbf{21.43} & \textbf{26.45} & 0.1712 & \textbf{5.887} \\
    \texttt{[0-1000]}   &  19.77 & 22.72 & -0.4529 & 5.111 \\
    \texttt{[250-750]} & 21.19 & 26.23 & \textbf{0.2658} & 5.581 \\
    \texttt{[500-750]} & 20.43 & 24.91 & -0.1553 & 5.582 \\
    \texttt{[250-500]} & 20.60 & 25.60 & 0.1037 & 5.336 \\
    \hline
\end{tabular}  
}
\end{minipage}
\begin{minipage}[t]{0.33\linewidth}
\setlength{\tabcolsep}{1.08mm}
\centering
\footnotesize
\caption{{Comparing win-lose pair choices. Choosing images of the highest and lowest quality is generally better than random selection. Note both strategies allow win-lose preference to align with aesthetic preference.}}
\label{tab:win_lose_choice}
{
\begin{tabular}{l|cccc}  
\hline
  win-lose sample  &  P-S & HPSV2 & I-R & AE\\  
    \hline  
    best \& worst    & \textbf{21.43} & 26.45 & \textbf{0.1712} & \textbf{5.887} \\
    random   &  21.21 & \textbf{26.51} & 0.1656 & 5.796 \\
    \hline
\end{tabular}  
}
\end{minipage}
\end{table*}

\subsection{Main Evaluation on Aesthetic Alignment}
\textbf{Automatic metrics.} Using the four automatic scores (Section \ref{sec:setup}), we compare SPO with Diffusion-DPO, D3PO, etc.,   
in Table~\ref{tab:prompt_alignment_matrix_sdxl} and Table~\ref{tab:auto_eval_matrix_sd15} based on SDXL and SD-1.5, respectively. {Note that ImageReward, PickScore, and HPSV2 assess overall image quality including aesthetics.} 

We have two observations. First, compared with vanilla SDXL and SD-1.5, the DPO methods yield consistent improvement in these metrics, demonstrating their effectiveness. 
Second, we observe that \textbf{SPO yields the best scores across the four metrics for both SDXL and SD-1.5}. For example, for SDXL, we achieve 23.06, 31.80, 1.0803, and 6.364 in PickScore, HPSV2, ImageReward, and Aesthetics, respectively. The improvement over Diffusion-DPO is +0.42, +2.49, +0.1367, and +0.349 on the four metrics, respectively. This again demonstrates the effectiveness of SPO in aesthetic improvement. 

\textbf{User studies.} We compare SPO with Diffusion-DPO and Vanilla SDXL through user studies. 
We ask annotators to compare three aspects: general preference, visual appeal, and prompt alignment 
and summarize the winning percentage in Fig.~\ref{fig:win_rate}. Results indicate that SPO has consistently more winning votes from users in general preference and visual appeal, and it is apparent that its visual appeal has a greater winning margin. For example, in general preference alignment, {SPO wins 52.8\% of all cases}, while in visual appeal, it wins 58.27\% of the cases. Similar observation can be made in SPO vs. vanilla SDXL.

\textbf{Qualitative comparison.} We show sample images in Fig.~\ref{fig:qualitative_comp}. Images generated by SPO are more visually appealing than those generated by SDXL and Diffusion-DPO. 

\subsection{Evaluation on Image-Text Alignment}
\textbf{SPO slightly improves vanilla models.} From Table \ref{tab:prompt_alignment_matrix_sdxl} and Table \ref{tab:auto_eval_matrix_sd15}, PickScore metric indicates some improvement over SDXL and SD-1.5. User studies (Fig. \ref{fig:win_rate}), on prompt alignment has similar results. 
{On GenEval, Table \ref{tab:geneval_results} shows SPO improves prompt alignment score by 1.77\% over SDXL.} 

\textbf{SPO has mixed results compared with Diffusion-DPO.} Table \ref{tab:prompt_alignment_matrix_sdxl} and Table \ref{tab:auto_eval_matrix_sd15} demonstrate improvement over Diffusion-DPO, and yet our user studies support a performance tie or SPO's slight win in prompt alignment. Further, GenEval results indicate that SPO is not as good as Diffusion-DPO in improving image-text alignment, which we tend to agree with, because the SPO design does not fully capture layout changes. But an interesting insight is that image-text alignment and aesthetics are probably often blended. So while SPO is designed for aesthetic alignment, it still yields some prompt alignment improvement, but the improvement is not as much as Diffusion-DPO.

\subsection{Further Analysis}\label{sec:ablation}
If not specified, we use SD-1.5 and Pick-a-Pic \emph{val.} set.

\begin{figure*}[t]
\begin{minipage}[t]{1\linewidth}
\centering
\includegraphics[width=1\textwidth]{imgs/typoclip_sdxl.pdf}
\caption{
{
Qualitative comparison between Glyph-ByT5-SDXL and Glyph-ByT5-SDXL + SPO in graphic design image generation. SPO consistently improves image aesthetics by creating nuanced textures and vibrant colors without sacrificing image content accuracy.}
}
\label{fig:glyphbyt5_comp}
\end{minipage}
\end{figure*}

\textbf{Effectiveness of step-aware preference model (SPM).} An important design of SPM is the addition of timestep conditioning. To verify its usefulness, we remove the timestep conditioning in SPM (SPM w/o step). A second variant is to simply use the PickScore model \cite{kirstain2024pick} which has no time condition and is trained on clean images only.  
Results in Table~\ref{tab:step_aware_preference_model} show that both variants lead to performance drop, validating the SPM design.

\textbf{Effectiveness of random selection for initializing next iteration.} At the end of each SPO iteration, we need to initialize the next iteration, where random selection is an option. Here we compare random selection with using the win sample $\boldsymbol{x}^w_{t-1}$ or the lose sample $\boldsymbol{x}^l_{t-1}$ for initialization. Results are presented in Table~\ref{tab:reinitialization}. We clearly find that random selection is better. If we only select $\boldsymbol{x}^w_{t-1}$ or $\boldsymbol{x}^l_{t-1}$, training becomes biased towards the intermediate samples that are more preferred or more dispreferred, respectively. This prevents the network from learning from more diverse trajectories, deteriorating generalization ability.

\textbf{Impact of number of candidates sampled at each denoising step.}
To find useful win-lose pairs, we obtain a set of candidates $\{\boldsymbol{x}_{t-1}^1, \boldsymbol{x}_{t-1}^2, \cdots, \boldsymbol{x}_{t-1}^k\}$ at each step, drawn from the conditional distribution $p_{\theta}\left(\boldsymbol{x}_{t-1} \mid \boldsymbol{x}_{t}, \boldsymbol{c}, t\right)$. 
{Table \ref{tab:num_of_sample_per_step_ablation} presents results by varying $k$.} We have two observations.

First, when increasing $k$, the discrepancy within sampled pairs becomes larger (but still small enough to only reflect image detail differences). The larger contrast between the preferred and dispreferred samples helps the model learn human preferences.
Second, when $k$ is too large, quality of the most dispreferred sample would be lower than average of samples generated by the base model. 
As a result, the ``push away'' effect of the dispreferred image is weakened, causing performance degradation. We choose $k=4$. 

\textbf{Impact of the number of inner steps $j$ in MSPO.} A larger $j$ allows the generated images to have higher diversity, which would be useful for strong diffusion models. In Table~\ref{tab:num_inner_step}, we study how $j$ impacts SPO. We observe that fine-tuning performance increases with $j$ in the beginning and then becomes saturated. When $j=1$, MSPO reduces to SPO. When $j$ goes to infinite, MSPO would essentially reduce to Diffusion-DPO because there will only be one step. From the table, we choose $j=4$.

\textbf{Impact of timestep range.} SPO is only applied to timestep range  [0-$\kappa$].  
Table \ref{tab:timestep_range} summarizes the results of applying SPO to various timestep ranges. We have the following observations. 

\textit{First}, discarding very large timesteps, \ie, [750-1000], yields better performance as these timesteps barely generate image details and are very noisy. \textit{Second}, if we remove [0-250] and only use [250-750], there is a considerate performance drop, indicating that [0-250] is useful. Similarly, [0-500] is also useful. \textit{Third}, if we compare [500-750], [250-500] and [0-250], we find that applying SPO to [250-500] achieves slightly better overall performance. We speculate [250-500] is a critical timestep range for SPO. Compared to larger timesteps, timesteps in [250-500] focus more on detail refinement. Moreover, compared to smaller timesteps, the denoising steps in [250-500] sample $\boldsymbol{x}_{t-1}$ with a sufficiently large variance to construct win-loss pairs for training. Based on these findings, we set $\kappa=750$ and apply SPO to timestep range [0-750].

\textbf{Comparing ways of choosing win-lose pairs.} In Table Table~\ref{tab:win_lose_choice}, we compare two ways. The proposed method uses a sample with highest quality and another sample with lowest quality. The other way is to randomly select two samples and use SPM to decide their win-lose preference. Results show that the proposed way is generally better. That said, both options allow for proper assessment of aesthetic preference because they ensure a sample pair comes from the same sample and has relatively small differences.

\textcolor{black}{\textbf{Computational cost.} We use \textcolor{black}{$4\times$ A100 GPUs}, which take $12$ and $29.5$ hours to fine-tune SD-1.5 and SDXL, respectively. We also spend $8$ and $29$ hours training SPM for SD-1.5 and SDXL, respectively. In comparison, the GPU hours used for fine-tuning SD-1.5 and SDXL using Diffusion-DPO are $384$ and $4,800$, respectively. As a result, the total training GPU hours of SPO-SD-1.5 and SPO-SDXL is about 20.8\% and 4.9\% of DPO-SD-1.5 and DPO-SDXL, respectively.} This significant efficiency gain is probably because of the SPO design, where the image details are properly highlighted for the network to learn.

\textbf{Generalization to Text Generation.} 
We verify the generalization of SPO by simply marrying the LoRA weights of SPO-SDXL to the Glyph-ByT5-SDXL model~\cite{liu2024glyph}, which specializes in design image generation. Qualitative examples are shown in Fig.~\ref{fig:glyphbyt5_comp}, where we observe that SPO consistently improves the visual appeal of Glyph-ByT5-SDXL images, \eg,   richer texture of the elephant, flower, robot, and beer mug, while preserving the text generation ability of Glyph-ByT5-SDXL.

\section{Conclusion}
This paper studies how to align diffusion models with human aesthetic preference. This problem is challenging for two reasons. First, existing two-trajectory methods exhibit large image discrepancies, preventing models from focusing on aesthetic nuances. Second, it is very non-trivial to collect aesthetic-only preference data, while existing datasets record generic preferences that may conflict with image aesthetic preference. To improve aesthetic alignment, we propose an aesthetic alignment solution that distills aesthetics from generic preference data. Specifically, the proposed step-by-step preference optimization method allows a pair of samples to originate from the same image, so their differences are relatively small after one or very few steps, which would reflect their image details or aesthetics. Our preference model captures these differences, enabling the model to improve towards generating better image details. In experiments, we demonstrate that SPO better aligns SD-1.5 and SDXL with human aesthetic preference compared with other DPO methods and is efficient to train.

{
    \small
    \bibliographystyle{ieeenat_fullname}
    \bibliography{main}
}

\clearpage
\setcounter{page}{1}
\setcounter{section}{0}
\maketitlesupplementary

\begin{figure}[t]
\begin{minipage}[t]{1.0\linewidth}
\centering
\begin{subfigure}[b]{1\textwidth}
\includegraphics[width=\linewidth]{imgs/supplementary_material/mismatch_traj1.pdf} 
\caption*{\textbf{Case A} Prompt: a brown purse abandoned on a green bench}
\end{subfigure}
\begin{subfigure}[b]{1\textwidth}
\includegraphics[width=\linewidth]{imgs/supplementary_material/mismatch_traj2.pdf} 
\caption*{\textbf{Case B} Prompt: a glazed donut with sprinkles, octane render, high quality, hyper realistic, vibrant colors, 4k, soft lighting}
\end{subfigure}
\caption{Image samples showing disagreement between generic preferences and aesthetic preference. These images are generated by SDXL. The win trajectories in both examples have inferior aesthetics / details, which are detailed in Section \ref{sec:examples} of the main text.}
\label{fig:disagreement_examples}
\end{minipage}
\end{figure}

\begin{figure}[t]
\begin{minipage}[t]{1\linewidth}
\centering
\includegraphics[width=1\textwidth]{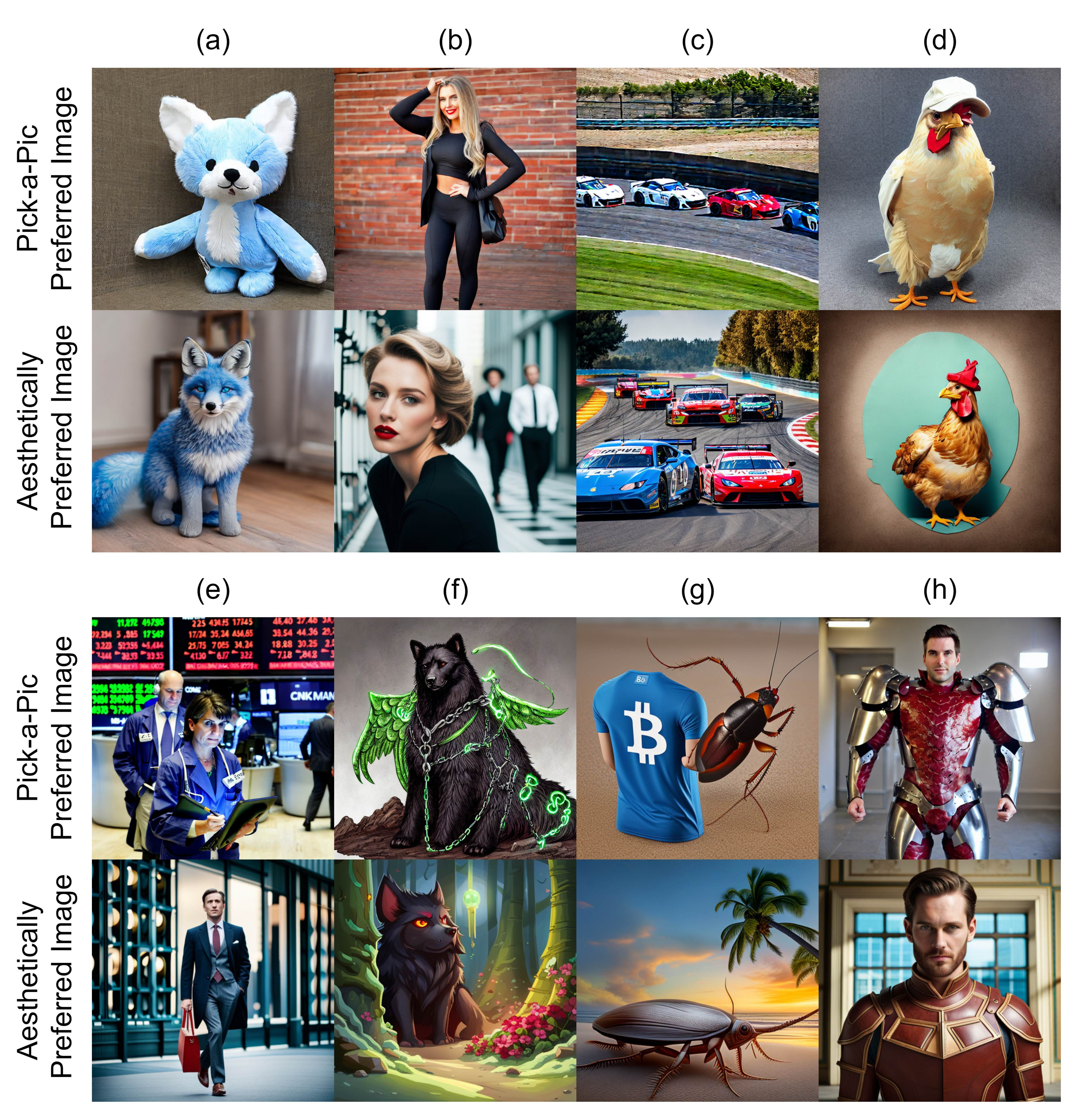}
\caption{More image samples showing disagreements between generic preferences and aesthetic preference in the Pick-a-Pic dataset 
(best viewed when zoomed in). These images were generated by various diffusion models. Prompt of (a): \textit{``a stuffed animal of a blue fox''}. Prompt of (b): \textit{``girl wearing red lipstick and black leggings''}. Prompt of (c): \textit{``4 cars racing''}. Prompt of (d): \textit{``gangsta clothed chicken''}. Prompt of (e): \textit{``Equity markets were mixed Monday''}. Prompt of (f): \textit{``A Kludde. A mythical monstrous black furry nocturnal dog with bear claws, green glistening scaled wings and glowing crimson eyes. Several heavy chains hang from its body and ankles''}. Prompt of (g): \textit{``**a portrait of a 3D cockroach, wearing a bitcoin shirt, in Hawaii, on the beach, hyper-realistic, ultra-detailed, photography, hyper-realistic, photo-realistic, ultra-photo-realistic, super-detailed, intricate details, 8K, surround lighting, HDR''}. Prompt of (h): \textit{``a suit of armour constructed from meat''}.}
\label{fig:mismatch_example2}
\end{minipage}
\end{figure}

\begin{figure*}[!htbp]
\begin{minipage}[t]{1\linewidth}
\centering
\includegraphics[width=0.68\textwidth]{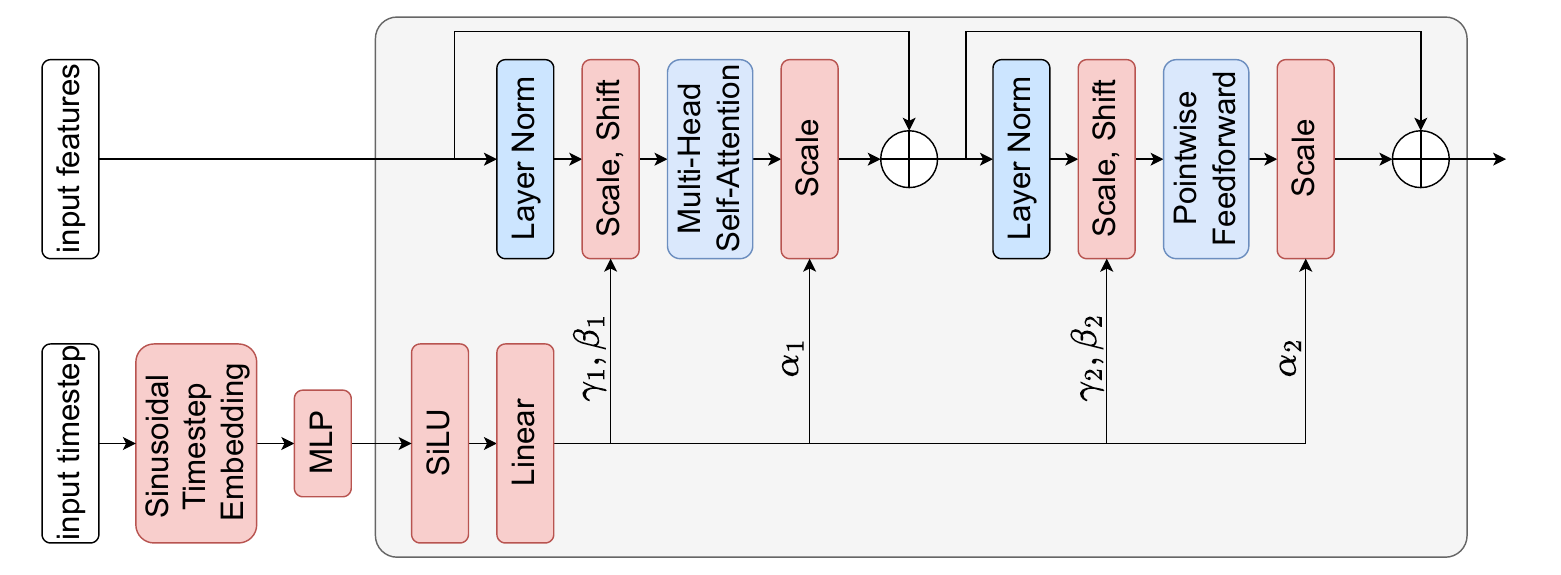}
\caption{Timestep-conditioned ViT block. The blue components are from the original ViT block and the red componets are newly added. $\oplus$ represents element-wise addition.}
\label{fig:t_cond_clip_v}
\end{minipage}
\end{figure*}


\section{More Examples of Disagreements Between Generic and Aesthetic Preferences}\label{sec:examples}
We show more examples of the disagreement in Fig. \ref{fig:disagreement_examples}. \redtext{Case A} of Fig. \ref{fig:disagreement_examples}: The output image of the upper trajectory is generally preferred because it aligns more closely with the prompt ``a brown purse abandoned \textcolor{red}{on a green bench}''. However, when considering aesthetic preference, the output image of the upper trajectory is dispreferred because it has some artifacts on the right side of the purse, while the output image of the lower trajectory has clean details. \redtext{Case B} of Fig. \ref{fig:disagreement_examples}: The output image of the upper trajectory is generally preferred over the one in the lower trajectory because it has the correct number of donuts as described in the prompt ``\textcolor{red}{a glazed donut} with sprinkles, octane render, high quality, hyper realistic, vibrant colors, 4k, soft lighting''. However, the color in the inner part of the donut of the upper output image is not consistent with the color in the background, making the output image of the lower trajectory aesthetically preferred.

We also examine the preference pairs from the Pick-a-Pic V1 dataset and showcase examples of disagreements between the Pick-a-Pic generic preference labels and aesthetic preference labels in Figure  \ref{fig:mismatch_example2}. These disagreements in the dataset hinder  model's improvement in aesthetics.

\section{Timestep-conditional CLIP Vision Encoder}
In this section, we present the implementation details of the timestep-conditional CLIP vision encoder introduced in Section \ref{sec:step_aware_preference_model} of the main text. The original CLIP vision encoder is based on a Vision Transformer (ViT) \cite{dosovitskiy2020image}. To incorporate timestep conditioning, we follow the approach in DiT \cite{peebles2023scalable}, employing time-conditional adaptive layernorm to inject timestep information into the vision encoder of CLIP. We replace all transformer blocks in the CLIP vision encoder with timestep-conditioned ViT blocks, the structure of which is illustrated in Figure \ref{fig:t_cond_clip_v}. The blue components in the figure represent the original Transformer block, while the red components denote the newly added components. We embed the input timestep as timestep embeddings using sinusoidal encoding, followed by an MLP with the structure \texttt{Linear-SiLU-Linear}. The same timestep embeddings are used as input for all timestep-conditioned ViT blocks.

A linear layer is employed to predict dimension-wise scaling parameters $\gamma$ and $\alpha$, as well as the dimension-wise shifting parameter $\beta$. Given input features $x$, the ``Scale, Shift'' operation modifies $x$ as $x = x \times (1 + \gamma) + \beta$, while the ``Scale'' operation adjusts $x$ as $x = x \times \alpha$. The ``Scale, Shift'' operation is applied directly after each layer normalization block, whereas the ``Scale'' operation is applied immediately after the multi-head self-attention and pointwise feedforward blocks, prior to the residual connections.

We initialize the step-aware preference model (SPM) with PickScore \cite{kirstain2024pick} weights. To preserve the pretrained knowledge, the weights of linear layers responsible for generating $\gamma$, $\beta$, and $\alpha$ are initialized such that $\gamma=0$, $\beta=0$, and $\alpha=1$, ensuring that the SPM's output matches the pretrained PickScore model's output at the start of training.


\section{Detailed Prompts}
We summarize the detailed text prompts used in Figure \ref{fig:qualitative_comp} of the main text in Table~\ref{tab:prompt_list}.

\begin{table*}[htbp]
\begin{minipage}[t]{1\linewidth}  
\centering  
\tablestyle{1pt}{1.2}  
\caption{  
Detailed prompts used for generated images in Figure \ref{fig:qualitative_comp} of the main text.}
\label{tab:prompt_list}
\resizebox{1.0\linewidth}{!}  
{  
\begin{tabular}{l|>{\centering\arraybackslash}m{16cm}}  
Image & Prompt \\  
\shline  
{Col1} & {Saturn rises on the horizon.} \\ \hline
{Col2} & {a watercolor painting of a super cute kitten wearing a hat of flowers} \\ \hline
{Col3} & {A galaxy-colored figurine floating over the sea at sunset, photorealistic.} \\ \hline
{Col4} & {fireclaw machine mecha animal beast robot of horizon forbidden west horizon zero dawn bioluminiscence, behance hd by jesper ejsing, by rhads, makoto shinkai and lois van baarle, ilya kuvshinov, rossdraws global illumination} \\ \hline
{Col5} & {A swirling, multicolored portal emerges from the depths of an ocean of coffee, with waves of the rich liquid gently rippling outward. The portal engulfs a coffee cup, which serves as a gateway to a fantastical dimension. The surrounding digital art landscape reflects the colors of the portal, creating an alluring scene of endless possibilities.} \\ \hline
{Col6} & {A profile picture of an anime boy, half robot, brown hair} \\ \hline
{Col7} & {Detailed Portrait of a cute woman vibrant pixie hair by Yanjun Cheng and Hsiao-Ron Cheng and Ilya Kuvshinov, medium close up, portrait photography, rim lighting, realistic eyes, photorealism pastel, illustration} \\ \hline
{Col8} & {On the Mid-Autumn Festival, the bright full moon hangs in the night sky. A quaint pavilion is illuminated by dim lights, resembling a beautiful scenery in a painting. Camera type: close-up. Camera lens type: telephoto. Time of day: night. Style of lighting: bright. Film type: ancient style. HD.} \\ 
\end{tabular}  
}
\end{minipage}
\end{table*}

\section{More Sample Images Generated by SPO-SDXL}
\begin{figure*}[htbp]
\begin{minipage}[!t]{1\linewidth}
\centering
\includegraphics[width=1\textwidth]{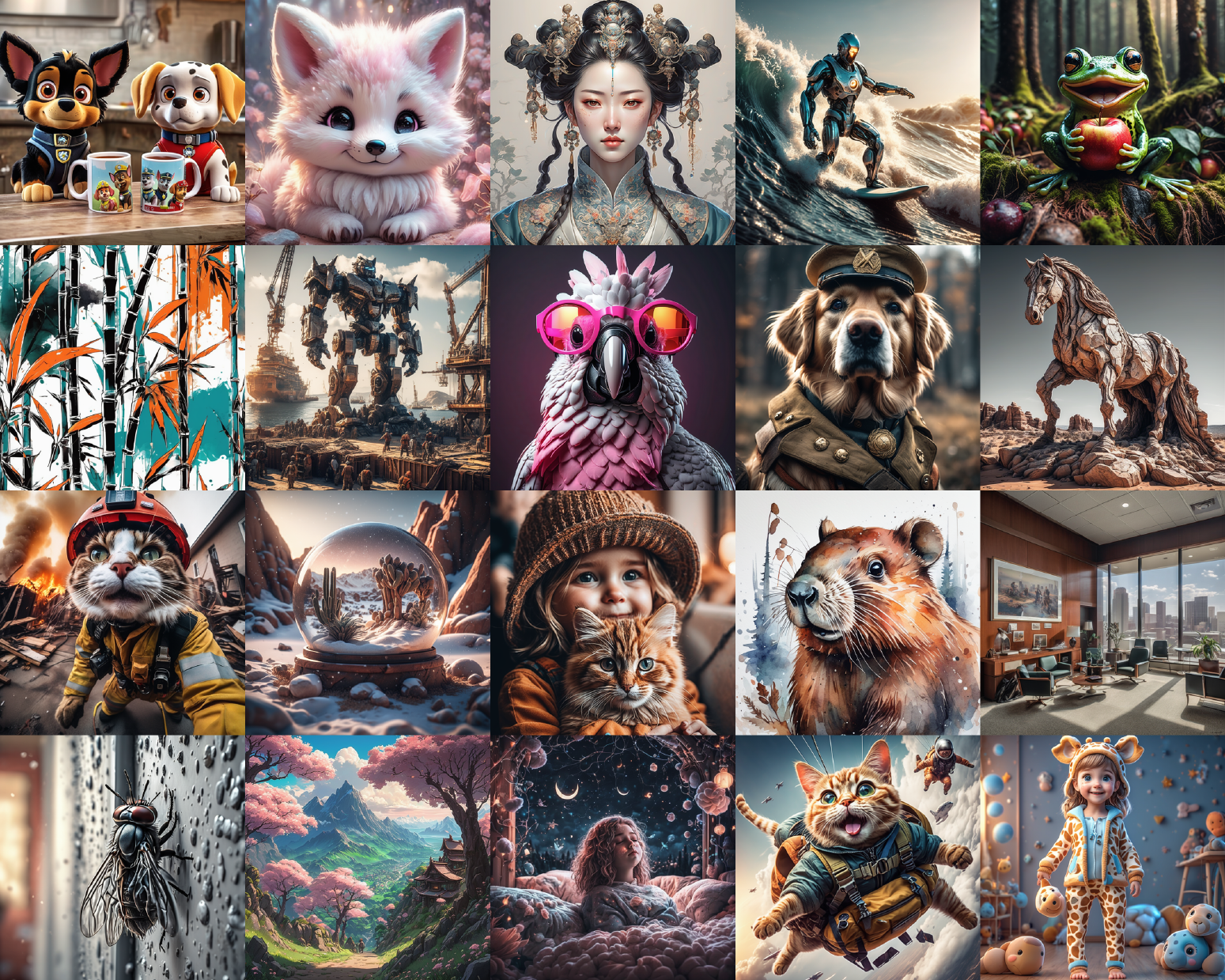}
\caption{Sample images generated by SPO-SDXL. With the SPO post-training, SPO-SDXL produces high-quality images that are visually attractive and stunning.}
\label{fig:more_sample_images}
\end{minipage}
\end{figure*}

Figure \ref{fig:more_sample_images} presents additional sample images generated by SPO-SDXL, accompanied by the corresponding text prompts listed in Table \ref{tab:more_sample_prompts}.

\begin{table*}[htbp]
\begin{minipage}[t]{1\linewidth}  
\centering  
\tablestyle{1pt}{1.2}  
\caption{  
Detailed prompts used for generated images in Figure \ref{fig:more_sample_images}.}
\label{tab:more_sample_prompts}
\resizebox{1.0\linewidth}{!}  
{  
\begin{tabular}{l|>{\centering\arraybackslash}m{16cm}}  
Image & Prompt \\  
\shline  
{Row 1, Col1} & {paw patrol. "This is some serious gourmet". 2 dogs holding mugs.} \\ \hline

{Row 1, Col2} & {little tiny cub beautiful light color White fox soft fur kawaii chibi Walt Disney style, beautiful smiley face and beautiful eyes sweet and smiling features, snuggled in its soft and soft pastel pink cover, magical light background, style Thomas kinkade Nadja Baxter Anne Stokes Nancy Noel realistic} \\ \hline

{Row 1, Col3} & {Full Portrait of Consort Chunhui by Giuseppe Castiglione, symmetrical face, ancient Chinese painting, single face, insanely detailed and intricate, beautiful, elegant, artstation, character concept in the style illustration by Miho Hirano, Giuseppe Castiglione --ar 9:16} \\ \hline

{Row 1, Col4} & {Surfer robot dude in the crest of a wave, cinematic, sunny, --ar 16:9} \\ \hline

{Row 1, Col5} & {a photo of a frog holding an apple while smiling in the forest} \\ \hline

{Row 2, Col1} & {185764, ink art, Calligraphy, bamboo plant :: orange, teal, white, black --ar 2:3 --uplight} \\ \hline

{Row 2, Col2} & {large battle Mecha helping with the construction of the Colossus of Rhodos standing above the entry of a harbor, hundreds of ancient workers on scaffolding surrounding the colossus, ancient culture, sunny weather, matte painting, highly detailed, cgsociety, hyperrealistic, --no dof, --ar 16:9} \\ \hline

{Row 2, Col3} & {A 3D Rendering of a cockatoo wearing sunglasses. The sunglasses have a deep black frame with bright pink lenses. Fashion photography, volumetric lighting, CG rendering, } \\ \hline

{Row 2, Col4} & {a golden retriever dressed like a General in the north army of the American Civil war. Portrait style, looking proud detailed  8k  realistic  super realistic  Ultra HD  cinematography  photorealistic  epic composition Unreal Engine  Cinematic  Color Grading  portrait Photography  UltraWide Angle  Depth of Field  hyperdetailed  beautifully colorcoded  insane details  intricate details  beautifully color graded  Unreal Engine  Editorial Photography  Photography  Photoshoot  DOF  Tilt Blur  White Balance  32k  SuperResolution  Megapixel  ProPhoto RGB  VR  Halfrear Lighting  Backlight  Natural Lighting  Incandescent  Optical Fiber  Moody Lighting  Cinematic Lighting  Studio Lighting  Soft Lighting  Volumetric  ContreJour  Beautiful Lighting  Accent Lighting  Global Illumination  Screen Space Global Illumination  Ray Tracing  Optics  Scattering  Glowing  Shadows  Rough  Shimmering  Ray Tracing Reflections  Lumen Reflections  Screen Space Reflections  Diffraction Grading  Chromatic Aberration  GB Displacement  Scan Lines  Ray Traced  Ray Tracing Ambient Occlusion  AntiAliasing  FKAA  TXAA  RTX  SSAO  Shaders} \\ \hline

{Row 2, Col5} & {A rock formation in the shape of a horse, insanely detailed,} \\ \hline

{Row 3, Col1} & {a gopro snapshot of an anthropomorphic cat dressed as a firefighter putting out a building fire} \\ \hline

{Row 3, Col2} & {a desert in a snowglobe, 4k, octane render :: cinematic --ar 2048:858} \\ \hline

{Row 3, Col3} & {cat, cute, child, hat} \\ \hline

{Row 3, Col4} & {watercolour beaver with tale, white background} \\ \hline

{Row 3, Col5} & {corporate office ralph goings -- aspect 3:2} \\ \hline

{Row 4, Col1} & {there once was a fly on the wall, I wonder, why didn’t it fall, Because its feet stuck, Or was it just luck, Or does gravity miss things so small, high realistic, high detailed, high contrast, unreal render --ar 16:9} \\ \hline

{Row 4, Col2} & {lush landscape with mountains with cherry trees by Miyazaki Nausicaa Ghibli, 王様ランキング, ranking of kings, spirited away, breath of the wild style, epic composition, clean  --w 1024 --h 1792 --no people} \\ \hline

{Row 4, Col3} & {what i dream when i close my eyes to sleep} \\ \hline

{Row 4, Col4} & {cute cat jumped off plane in parachute, exaggerated expression, photo realism, side angle, epic drama} \\ \hline

{Row 4, Col5} & {Full body, a Super cute little girl, wearing cute little giraffe pajamas, Smile and look ahead, ultra detailed sky blue eyes, 8k bright front lighting, fine luster, ultra detail, hyper  detailed 3D rendering s750, } \\
\end{tabular}  
}
\end{minipage}
\end{table*}  

\end{CJK}
\end{document}